\documentclass[sn-mathphys-num]{sn-jnl}

\usepackage{graphicx}%
\usepackage{multirow}%
\usepackage{amsmath,amssymb,amsfonts}%
\usepackage{amsthm}%
\usepackage{mathrsfs}%
\usepackage[title]{appendix}%
\usepackage{xcolor}%
\usepackage{textcomp}%
\usepackage{manyfoot}%
\usepackage{booktabs}%
\usepackage{algorithm}%
\usepackage{algorithmicx}%
\usepackage{algpseudocode}%
\usepackage{listings}%

\usepackage{anyfontsize}
\usepackage{ragged2e} 
\usepackage{multirow}
\usepackage{array}

\makeatletter
\renewcommand\paragraph{\@startsection{paragraph}{4}{\z@}%
                                    {3.25ex \@plus1ex \@minus.2ex}%
                                    {-1em}%
                                    {\normalfont\normalsize\bfseries}}
\makeatother
\begin{document}

\title[Penalty Learning for Optimal Partitioning using Multilayer Perceptron]{Penalty Learning for Optimal Partitioning using Multilayer Perceptron}

\author*[1]{\fnm{Tung} \sur{L Nguyen}}\email{tln229@nau.edu}
\author[2]{\fnm{Toby} \sur{Dylan Hocking}}\email{toby.dylan.hocking@usherbrooke.ca}

\affil*[1]{\orgdiv{School of Informatics, Computing, and Cyber Systems}, \orgname{Northern Arizona University}, \orgaddress{\street{S San Francisco}, \city{Flagstaff}, \postcode{86011}, \state{Arizona}, \country{USA}}}
\affil[2]{\orgdiv{Département d'informatique}, \orgname{Université de Sherbrooke}, \orgaddress{\street{Sherbrooke QC J1K 2R1}, \city{Quebec}, \country{Canada}}}

\abstract{Changepoint detection is a technique used to identify significant shifts in sequences and is widely used in fields such as finance, genomics, and medicine.
To identify the changepoints, dynamic programming (DP) algorithms, particularly Optimal Partitioning (OP) family, are widely used.
To control the changepoints count, these algorithms use a fixed penalty to penalize the changepoints presence.
To predict the optimal value of that penalty, existing methods used simple models such as linear or tree-based, which may limit predictive performance.
To address this issue, this study proposes using a multilayer perceptron (MLP) with a ReLU activation function to predict the penalty.
The proposed model generates continuous predictions -- as opposed to the stepwise ones in tree-based models -- and handles non-linearity better than linear models.
Experiments on large benchmark genomic datasets demonstrate that the proposed model improves accuracy and F1 score compared to existing models.}

\keywords{changepoint detection, partitioning, segmentation, penalty learning, multilayer perceptron}

\maketitle

\begin{figure}[!tb]
\includegraphics[width=0.99\textwidth]{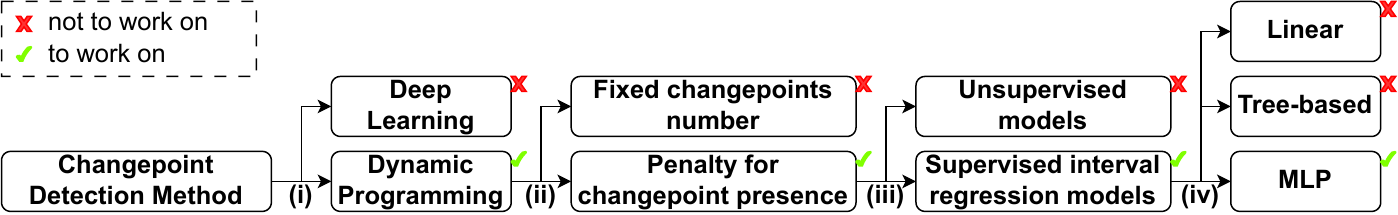}
\caption{Literature review diagram of the changepoint detection process in this study: (i) Selecting a detection method, (ii) Choosing a regularization method to control the number of changepoints, (iii) Selecting a model type to predict the penalty, and (iv) Choosing the model architecture.}
\label{fig:literature}
\end{figure}

\section{Introduction}\label{sec:intro}

Changepoint detection (or other words, partitioning and segmentation) is essential in many real-world applications, identifying significant data shifts in finance \citep{app:finance}, healthcare \citep{cancer}, network security \citep{app:network}, environmental monitoring \citep{app:climate}, and more.

Detecting changepoints involves selecting the detection method and tuning hyperparameters to detect the changepoint locations.
Each literature review paragraph below is numbered corresponding to the steps illustrated in Figure \ref{fig:literature}.

\paragraph{(i) Changepoint detection methods review}
Many real-life time series data are univariate.
Several deep learning (DL)-based changepoint detection methods exist \citep{review:londschien2023randomforest, review:lee2023, review:li24automatic, review:ermshaus2023clasp}, these approaches generalize classical methods \citep{review:cusum, review:james1987tests} by applying sliding classification windows built using either multilayer perceptron (MLP) or k-nearest neighbors (KNN). 
DL-based methods are fairly easy to understand and implement, making them accessible to a wide audience.
But these methods require tuning numerous hyperparameters, making them overly complex for univariate data.
In contrast, dynamic programming (DP) algorithms provides an ideal solution in this context.
\textit{This study selects DP to enhance its changepoint detection performance.}

\paragraph{(ii) Regularization for Changepoints Count in DP Algorithms}
DP algorithms detects changepoints using only one hyperparameter (contrast to numerous hyperparameters in DL-based methods) to control changepoints count, either (a) a fixed changepoints count or (b) a penalty to penalize changepoint presence.
Approach (a), used in \citep{auger1989algorithms, bai2003computation, killick2012optimal}, is effective but relies on often unknown changepoints count.
Approach (b), with the most popular methods being Optimal Partitioning (OP) \citep{jackson:05} and its variants, including Pruned Exact Linear Time (PELT) \citep{killick2012optimal}, Functional Pruning Optimal Partitioning (FPOP) \citep{Maidstone:16}, and Labeled Optimal Partitioning (LOP) \citep{hocking:20}.
PELT and FPOP improves efficiency by pruning technique whilst producing the same solution as OP. 
LOP extends OP by incorporating predefined changepoint labels and defaults to OP when labels are absent.
\textit{This study selects approach (b) to enhance changepoint detection performance as it offers greater flexibility.}

\paragraph{The role of the OP penalty}
A higher penalty imposes stricter penalties, reducing the changepoints count (see Figure \ref{fig:lambda}).
For a better understanding, see Appendix section \ref{OP optimization}.
% While these algorithms are effective, the fixed nature of OP penalty parameter raises the need for adaptive methods.
Given the suitability of OP family algorithms for changepoint detection in univariate sequences and the role of the OP penalty, \textit{this study focuses on developing a model to predict the optimal OP penalty}.

\begin{figure}[!tb]
\centering
\includegraphics[width=0.99\textwidth]{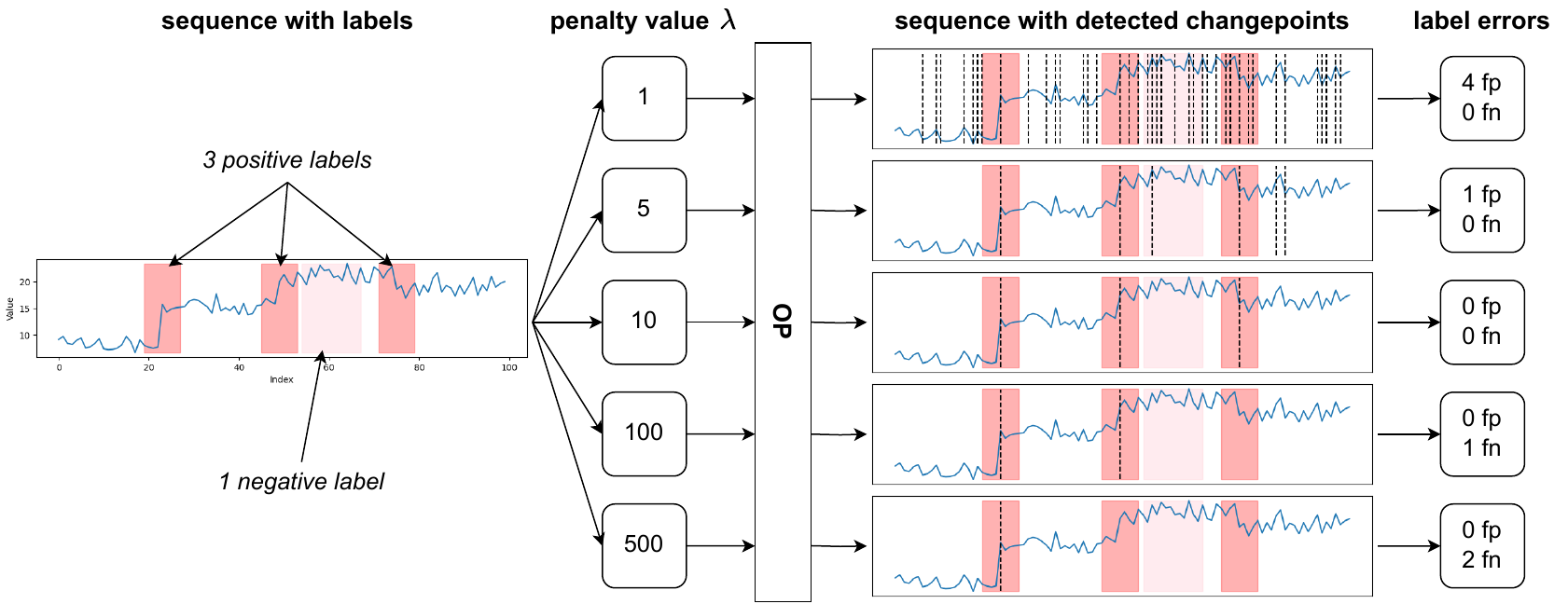}
\caption{Example: how \(\lambda\) value influences changepoints count. The labeled sequence has 3 positive labels (each with one changepoint) and 1 negative label (with no changepoints). Label errors are classified as false positives (fp), occurring when multiple changepoints are detected in a positive label or any changepoint is detected in a negative label, and false negatives (fn), occurring when no changepoint is detected in a positive label. In this example, $\lambda = 10$ is optimal.}
\label{fig:lambda}
\end{figure}

\paragraph{(iii) Related Work review} Various methods exist for predicting the OP penalty, including unsupervised \citep{Schwarz:78, akaike1974new, Lavelle:05} and supervised ones \citep{truong17, rigaill:13, MMIT, barnwal2022survival}. Details are in Section \ref{sec:related_work}.

\paragraph{(iv) Gap in the Literature and Contributions}  
Existing models are constrained by simplistic architectures, such as linear or tree-based, which motivated us to propose a novel approach. 
This study proposes MLP with activation function ReLU for OP penalty prediction. 
This is a generalized version of a linear model and offers continuous nonlinear predictions compared to tree-based models. 
Experiments conducted on three large genome datasets demonstrate that the proposed model outperforms previous work in accuracy and F1 score.
For simplicity, we refer to the OP penalty as $\lambda$ hereafter.

% \paragraph{Paper structure} This study is structured into five main sections.
% Section \ref{sec:intro} introduces the importance of \(\lambda\) prediction.
% In Section \ref{sec:related_work}, problem setting and previous methods on $\lambda$ prediction is reviewed. 
% Section \ref{sec:novelty_contribution} details our method, highlighting innovations that overcome the limitations of previous approaches. 
% Section \ref{sec:experiments} shows the results obtained from applying our proposed method are presented and analyzed comprehensively. 
% Section \ref{sec:conclusion} comprises the discussion and conclusion of the study.

\section{Related Work}\label{sec:related_work}
The related work and the proposed method are interval regression models. 
Before discussing the related work, the next part provides study supervision and explains how they are interval regression models instead of point regression.

\subsection{Problem setting}
\paragraph{Study supervision} Each sequence has \textit{labels} indicating regions with an expected changepoints count.
A \textit{label error} occurs when detected changepoints counts deviate from expectations. 
The optimal $\lambda$ for a labeled sequence minimizes label error.
For example in Figure \ref{fig:lambda}, the sequence has four labels, \(\lambda = 10\) is considered optimal.

\begin{figure}[!tb]
\centering
\includegraphics[width=0.99\textwidth]{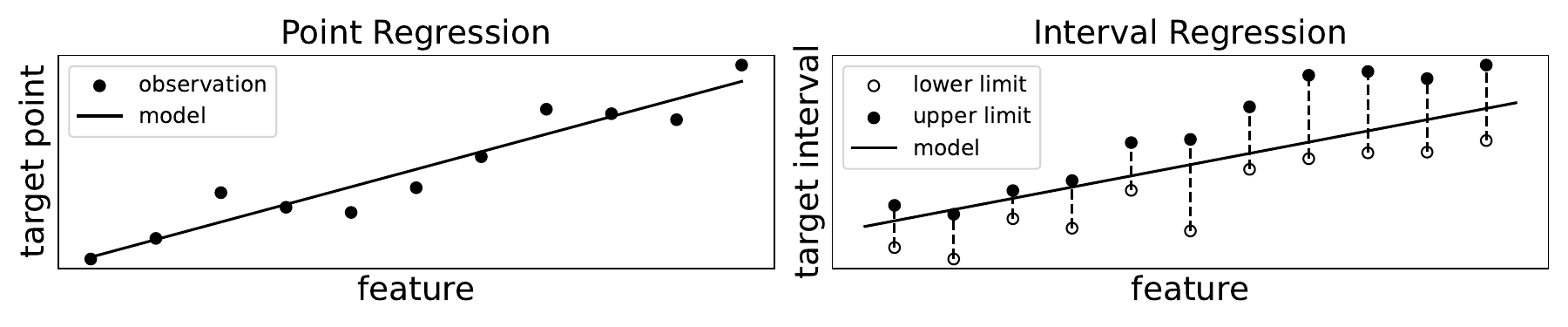}
\caption{Point Regression vs Interval Regression:
In point regression, the model line tries to pass close to all the targets. 
In Interval Regression, the model line aims to intersect all the target intervals.}
\label{fig:interval_regression}
% \includegraphics[width=0.99\textwidth]{fig.loss_function.pdf}
% \caption{The loss value of two loss functions is determined by comparing the prediction to the target.
% In the left plot, the SE yields a value of 0 when the prediction $\hat{y}$ reaches the target point $y$. The right plot illustrates the Squared Hinge Error (this loss function is utilized to predict values falling within a target interval rather than a single target point), where the loss value reaches 0 when the prediction $\hat{y}$ falls within the interval $[y_l + \epsilon, y_u - \epsilon]$.}
% \label{fig:loss_func}
\end{figure}

\begin{figure}[!tb]
\centering
\includegraphics[width=0.88\textwidth]{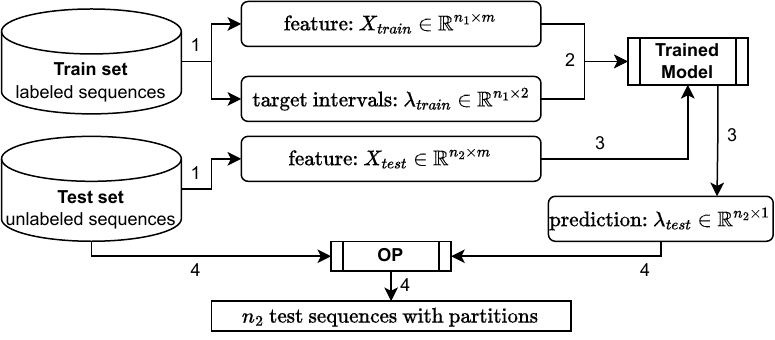}
\caption{The changepoint detection process: a train set with $n_1$ labeled sequences and a test set with $n_2$ unlabeled sequences. Step 1: Extract $m$ features from both sets and get a target interval from each labeled sequence in the train set. Step 2: Train the interval regression model using the features and target intervals from the train set. Step 3: Use the trained model to predict $\lambda$ for each test sequence. Step 4: For each unlabeled sequence in the test set, apply OP using the predicted $\lambda$ obtained from the trained model.}
\label{fig:problem_setting}
\end{figure}

\paragraph{The $\lambda$ prediction models are interval regression} The model input is a features set from sequences (length, variance, etc).
% \citep{hocking:20} uses a single feature, \(\log N\) and \citep{rigaill:13} employs a feature vector \([\log N, \sigma]^T\), where \(N\) and \(\sigma\) represents the sequence length and variance, respectively. Furthermore, \citep{rigaill:13} includes various statistical features and transformations (such as square, square root, absolute, log, loglog, etc.) in the sequence feature vector, creating a larger and more comprehensive feature vector.
The model target is the optimal \(\lambda\) interval (not one single optimal value), or \textit{target interval} (see \citep{rigaill:13} of how to get the optimal $\lambda$ interval for a labeled sequence). 
The predicted $\lambda$ should fall within the target interval, see Figure \ref{fig:interval_regression}. This is an \textit{interval regression} problem, with four types of intervals $[y_l,y_u]$: uncensored $( -\infty < y_l = y_u < \infty )$, interval-censored $( -\infty < y_l < y_u < \infty )$, left-censored $( -\infty = y_l < y_u < \infty )$, and right-censored $( -\infty < y_l < y_u = \infty )$.
The related work or proposed method is step 2 of the changepoint detection process in Figure \ref{fig:problem_setting}.

% \paragraph{Changepoint Detection Process Summary} 
% For each labeled sequence in the train set, a feature vector is generated and the target interval is determined based on its labels. During model training, the feature vector serves as input to predict \(\lambda\), aiming to keep it within the target interval. The predicted \(\lambda\) is then used in OP to detect changepoints in the new unlabeled sequences. A step-by-step process is illustrated in Figure \ref{fig:problem_setting}.

\subsection{Related work on $\lambda$ prediction}
\paragraph{Noteworthy} Adaptive Linear Penalty INference (ALPIN) \citep{truong17} is a method for a special case of this supervision, where every point in the sequence is a label, equivalent to label regions having length 1.
Since this study uses a more relaxed supervision with varying label region lengths, ALPIN is not applicable.
Accelerated Failure Time (AFT) models, a class of models for censored outcomes, can be considered. 
While changepoint detection datasets may include various interval types, traditional AFT models \citep{wei1992accelerated, cai2009regularized, huang2006regularized, quinlan1986induction, rf, svm} are limited to uncensored and right-censored intervals, making them inapplicable.

\paragraph{Unsupervised}
The Bayesian Information Criterion (BIC) \citep{Schwarz:78} sets \(\lambda = \log N\), where \(N\) is the sequence length, while Akaike's Information Criterion (AIC) \citep{akaike1974new} selects \(\lambda = 2p\), with \(p\) is a feature (e.g, variance). \citet{Lavelle:05} does not specify a particular form but computes \(\lambda\) based on sequence length and variance.

\paragraph{Supervised}
The AFT in the XGBoost model \citep{barnwal2022survival}, despite being an AFT model, can handle all non-negative target intervals, making it suitable for this study.

Since \(\lambda\) is non-negative, to eliminate this constraint in models output, the following models predict \(\log \lambda\) instead of \(\lambda\): linear \citep{rigaill:13} and Maximum Margin Interval Trees (MMIT) \citep{MMIT}. 
Both are trained by using a specific loss function, the Hinge Error. 
Using the ReLU function \big(\(f(x) = \max(0,x)\)\big), Hinge Error between prediction $\hat{y}$ and target interval $[y_l, y_u]$ is expressed as:

\begin{equation}\label{loss_function}
l(\hat{y}, [y_l, y_u]) = \big( \text{ReLU}(y_l - \hat{y} + \epsilon)\big)^p + \big(\text{ReLU}(\hat{y} - y_{u} + \epsilon) \big)^p
\end{equation}

\noindent Here, $\epsilon \geq 0$ is the margin length (with \citep{rigaill:13} using $\epsilon = 1$ and \citep{MMIT} treating $\epsilon$ as a hyperparameter), and $p$ (1 or 2) defines the loss type.
It is clear that if the prediction $\hat{y}$ falls within the interval $[y_l + \epsilon, y_u - \epsilon]$, the Hinge Error value is 0.

\section{Novelty and Contribution}\label{sec:novelty_contribution}

\paragraph{Novelty: MLP for $\lambda$ prediction}
A MLP with fully connected architecture is used.
All hidden layers have the same number of neurons.
ReLU activation function is used in the input and hidden layer(s). This architecture makes the proposed model a more general form of a linear model and produces continuous nonlinear predictions compared to tree-based models like MMIT and AFT in XGBoost.

\paragraph{Additional Contribution: Feature Selection}
A meaningful feature distinguishes sequences from one another in the context of changepoint detection.
Features such as the mean, quartile values (min, Q1, Q2, Q3, max) and their transformations are not selected because these features are noise in the changepoint context.
For example, \([0,0,0,1,1,1]\) and \([1,1,1,2,2,2]\) are the same sequences.
Sequence length and variance alone cannot fully represent a sequence, since some sequences may have the same length and variance but differ significantly in this context, so we introduce two additional sequence-related features: \textit{value range} and \textit{sum of absolute differences}.
\begin{itemize}
\item \textbf{Value range:} This feature ($r$ in Table \ref{tab:models}) is defined as the difference between the maximum and minimum values in the sequence.
\item \textbf{Sum of absolute differences:} This feature ($s$ in Table \ref{tab:models}) is calculated as the sum of the absolute differences between two consecutive sequence points, denoted as $\sum_{i=1}^{N-1} |d_{i+1} - d_{i}|$, where $\textbf{d} = [d_1, d_2, \dots, d_N]$ is the sequence. This feature differentiates sequences with the same length, variance, or value range, such as [0,0,0,1,1,1] and [0,1,0,1,0,1], which are distinct in the context of changepoint detection.
\end{itemize}
Intuitively, as the value range or the sum of absolute differences increases, the sequence tends to show more fluctuations, indicating a need to increase $\lambda$.
See Figure \ref{fig:features_targets} in Appendix \ref{sup_figures} to see why these two features emerge as strong candidates.

\section{Experiments}\label{sec:experiments}

% \paragraph{Evaluation Metric}
% The evaluation metric is the accuracy rate and F1-score of the test set, calculated as the ratio of label errors to the labels count.

\paragraph{Features} Four sets of features are considered to evaluate their quality:
\begin{itemize} 
\item \textbf{1 feature}: Sequence length, as used in \cite{hocking:20} and \cite{Schwarz:78}. 
\item \textbf{2 features}: Sequence length and variance, following \cite{rigaill:13}. 
\item \textbf{4 features}: Sequence length, variance, value range, sum of absolute differences. 
\item \textbf{All features}: According to \cite{rigaill:13} and \cite{MMIT}, up to 365 features are generated, with only those without of missing data, infinite values, or zero variance retained.
\end{itemize}

\begin{table}[!tb]
\centering
\caption{Datasets used in the study: Positive labels refer to regions in the sequence that contain changepoint(s), while negative labels indicate regions without any changepoints.}
\begin{tabular}{llllll}
\hline
\textbf{Dataset} & \textbf{Sequences} & \textbf{Pos. Labels} & \textbf{Neg. Labels} & \textbf{Source}                                                                                                   & \textbf{Used In}  \\ \hline
{Detailed}       & 3730                    & 964                  & 3396                 & \href{https://github.com/tdhock/neuroblastoma-data/tree/master/data/detailed}{GitHub}                             & \citep{rigaill:13} \\ \hline
{Systematic}     & 3418                    & 573                  & 2846                 & \href{https://github.com/tdhock/neuroblastoma-data/tree/master/data/systematic}{GitHub}                           & \citep{rigaill:13, MMIT} \\ \hline
{Epigenomic}     & 4913                    & 2185                 & 8195                 & \href{https://archive.ics.uci.edu/ml/machine-learning-databases/00439/peak-detection-data.tar.xz}{UCI Repo} & \citep{barnwal2022survival} \\ \hline
\end{tabular}
\label{tab:datasets}
\end{table}

\paragraph{Models Configuration}
The models list is in Table \ref{tab:models}, and the implementation packages for each are provided in Appendix \ref{sup_tables}, Table \ref{tab:packages}.
All models employed 5-fold cross-validation to select the optimal hyperparameters.
\begin{itemize}
    \item \textbf{Linear}: L1 regularization is applied only to the large feature vector, with cross-validation used to determine the optimal L1 parameter value, starting at 0.001 and increasing by 1.2 until no features remain.

    \item \textbf{MMIT}:
The hyperparameters considered include a list for \texttt{max\_depth} values (0, 1, 5, 10, 20, Inf), \texttt{margin} (0, 1, 2) ($\epsilon$ from equation \ref{loss_function}), \texttt{loss\_type} (\texttt{hinge}, \texttt{square} -- equivalent to $p=1$ or $p=2$ in Equation \ref{loss_function}) and \texttt{min\_sample} (0, 1, 2, 4, 8, 16, 20).

    \item \textbf{AFT in XGBoost}: 
The hyperparameters considered include a list of \texttt{learning\_rate} values (0.001, 0.01, 0.1, 1.0), \texttt{max\_depth} (2, 3, 4, 5, 6, 7, 8, 9 10), \texttt{min\_child\_weight} (0.001, 0.1, 1.0, 10.0, 100.0), \texttt{reg\_alpha} (0.001, 0.01, 0.1, 1.0, 10.0, 100.0), \texttt{reg\_lambda} (0.001, 0.01, 0.1, 1.0, 10.0, 100.0), \texttt{aft\_loss\_distribution\_scale} (0.5, 0.8, 1.1, 1.4, 1.7, 2.0).
This setup is identical to that described in \citep{barnwal2022survival}.

    \item \textbf{MLP}: The hyperparameters considered include a list of \texttt{n\_hidden\_layer} (1, 2, 3, 4) and \texttt{hidden\_layer\_size} (2, 4, 8, 16, 32, 64, 128, 256, 512). We used the early stopping technique from \citep{earlystop} using a \texttt{patience} parameter, stopping training if the loss does not decrease after the specified patience iterations; in our study, we set the maximum iterations to 12,000 and the patience to 20. The model is trained using the Adam optimizer (with default PyTorch hyperparameters) to minimize the Squared Hinge Loss (as defined in Function \ref{loss_function} with $p=2$ and $\epsilon=1$).
\end{itemize}

\paragraph{Results} This study uses 3 large datasets, as detailed in Table \ref{tab:datasets}.
Each dataset is split into 6 similar size folds.
In each iteration, 1 fold is designated as the test set, and the remaining 5 folds form the train set, with the process yield 6 test accuracy rates (Figure \ref{fig:acc_comp}) and F1 scores (Figure \ref{fig:F1_comp}) for each method.

\begin{table}[!tb]
\centering
\caption{List of employed models}\label{tab:models}%
\begin{tabular}{@{}lllll@{}}
\toprule
\textbf{Model} & \textbf{Features} & \textbf{Model Type} & \textbf{Regularization} & \textbf{Citation} \\
\midrule
BIC.1 & $\log\log N$ & unsupervised & None & \cite{Schwarz:78} \\
\midrule
linear.1 & $\log\log N$ & \multirow{3}{*}{linear} & \multirow{3}{*}{None} & \multirow{3}{*}{\begin{tabular}[c]{@{}l@{}}\cite{hocking:20}\\ \cite{rigaill:13}\\ \cite{Lavelle:05}\end{tabular}} \\
linear.2 & $\log\log N, \log \sigma$ &  &  & \\
linear.4 & $\log\log N, \log \sigma, \log r, \log\log s$ &  &  & \\
\midrule
linear.all & all features &  & L1 & \cite{rigaill:13} \\
\midrule
mmit.1 & $N$ & \multirow{4}{*}{tree} & max depth & \multirow{4}{*}{\cite{MMIT}} \\
mmit.2 & $N, \sigma$ &  & min split sample & \\
mmit.4 & $N, \sigma, r, s$ &  & loss type & \\
mmit.all & all features &  & loss margin & \\
\midrule
 &  &  & distribution scale & \\
aft\_xgboost.1 & $N$ & \multirow{4}{*}{ensemble tree} & learning rate & \multirow{6}{*}{\cite{barnwal2022survival}} \\
aft\_xgboost.2 & $N, \sigma$ &  & max depth & \\
aft\_xgboost.4 & $N, \sigma, r, s$ &  & min child weight & \\
aft\_xgboost.all & all features &  & reg alpha & \\
 &  &  & reg lambda & \\
\midrule
mlp.1 & $\log\log N$ & \multirow{4}{*}{MLP} & hidden layers count & \multirow{4}{*}{proposed} \\
mlp.2 & $\log\log N, \log \sigma$ &  & neurons per layer & \\
mlp.4 & $\log\log N, \log \sigma, \log r, \log\log s$ &  & early stopping & \\
mlp.all & all features &  &  & \\
\botrule
\end{tabular}
\footnotetext{Notations: length $N$ --- variance $\sigma$ --- value range $r$ --- sum of absolute difference $s$}
\end{table}

% \paragraph{MLP Configurations}
% To implement each MLP model, a cross-validation (cv=5) was used on the train set to determine the optimal number of hidden layers and the number of neurons per hidden layer.
% This selection was based on achieving the highest accuracy rate on the corresponding validation set.
% Different MLP configurations were validated, including models with 1 to 4 hidden layers and
% all hidden layer of one model can have sizes of 2, 4, 8, 16, 32, 64, 128, 256 or 512 neurons.

% \paragraph{Number of iterations with Early Stopping}
% There is one early stopping technique, presented in \citep{earlystop}, controls the number of iterations. 
% It uses the ``patience" parameter to decide when to stop learning. 
% During neural network training, setting a large maximum number of iterations and specifying the value for patience parameter allows for monitoring the training process. 
% If the loss value from the train set does not decrease after the specified number of patience iterations, the training process is halted. 
% This technique helps prevent overfitting.
% In our study, we set the maximum number of iterations to 12000 and patience parameter value to 20 iterations.

\section{Discussion }\label{sec:discussion}
\subsection{Performance}
\begin{figure}[!tb]
\includegraphics[width=1.0\textwidth]{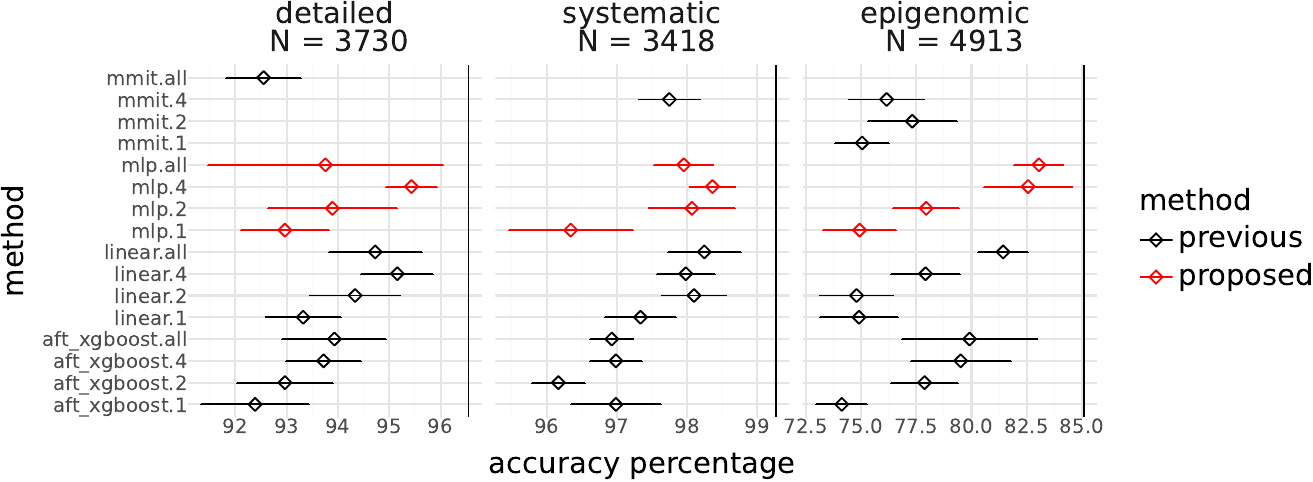}
\caption{Mean test accuracy and ±1 standard deviation across 6 folds for each method. To enhance clarity in the comparison visualization, method BIC and some small accuracies have been omitted.}
\label{fig:acc_comp}
\end{figure}

\begin{figure}[!tb]
\includegraphics[width=1.0\textwidth]{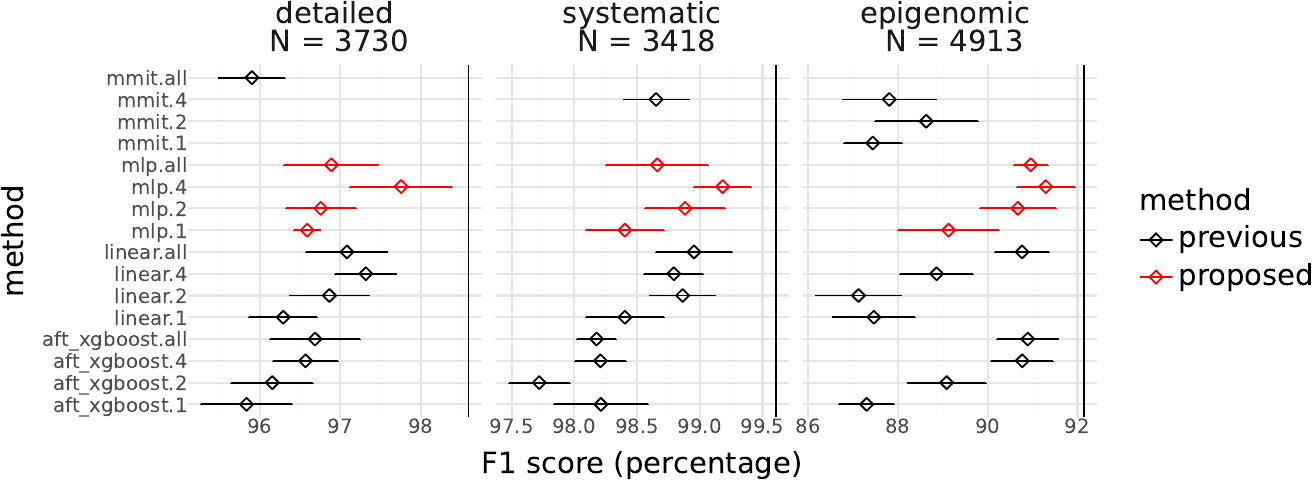}
\caption{Mean test F1 score and ±1 standard deviation across 6 folds for each method. To enhance clarity in the comparison visualization, method BIC and some small F1 score have been omitted.}
\label{fig:F1_comp}
\end{figure}

\paragraph{Overall Comparison} Figures \ref{fig:acc_comp} and \ref{fig:F1_comp} illustrate that the proposed model, using four features, performs slightly better than previous models across all datasets.

\paragraph{Comparison between previous methods} In many scenarios, with the same features set, tree-based models do not outperform linear models. This observation is consistent with results reported in \citep{MMIT} for the dataset Systematic, and \citep{barnwal2022survival} for the dataset Epigenomic (this study used 10 sub-sets from the large set Epigenomic). This lack of improvement is due to observable linear relationships between features and target intervals, as shown in Figure \ref{fig:features_targets}, Appendix \ref{sup_figures}.

\paragraph{Proposed Model vs. Linear} As shown in Figures \ref{fig:acc_comp} and \ref{fig:F1_comp}, performance generally improves with an increasing number of features. For the same feature set, the proposed model -- a piecewise linear model -- usually outperforms the linear model, as expected.

\paragraph{Consistency of the Proposed Model} Figures \ref{fig:acc_comp} and \ref{fig:F1_comp} show that MLP models generally exhibit a larger confidence interval (accuracy and F1 score) compared to other models, likely due to the complexity of the proposed model's architecture.

\subsection{Factors Affecting Performance}

\paragraph{Features} The experiments show that performance improves as the number of features increases (from 1 to 2 to 4, excluding the case of all features), regardless of whether linear, tree-based, or MLP models are used.
The reason lies in the better quality of the larger feature set. 
If we closely examine the feature sequence length in Figure \ref{fig:features_targets}, Appendix \ref{sup_figures}, the relationship between sequence length and the target interval of $\lambda$ is not entirely clear. 
This lack of clarity could explain why models relying solely on this feature exhibit worse performance compared to others.
Noticeably, nonlinear models don't have better performance when employing all features compared to when using only 4 features.
So selecting only 4 features proves highly beneficial, enhancing model performance while also potentially reducing training time.

\paragraph{MLP Configurations} Configuring MLPs appropriately (refer to Figure \ref{fig:mlp} in Appendix \ref{sup_figures}) leads to a performance improvement compared to previous models. 
Since linear models perform well on these datasets, as noted in \citep{MMIT} and \citep{barnwal2022survival}, the proposed model — a piecewise linear model — also achieves fairly good performance with just a single hidden layer and a small number of neurons (4 or 8).

\subsection{Limitations of this study}

% \paragraph{May not be applicable to other sequence datasets} Intuitively, features such as variance, value range, and sum of absolute difference play a role in predicting this penalty problem for these particular benchmark datasets. However, these features may not be relevant in other types of sequence datasets.

\paragraph{Overlooking other useful features} The number of potential features that can be extracted from a sequence is theoretically infinite. For instance, features like skewness, kurtosis, or the count of repeated values were not considered here—highlighting that many other useful features might easily be overlooked.

\paragraph{Absence of an inherent feature selection mechanism
} Unlike L1-regularized linear models and tree-based methods, MLPs do not have a built-in feature selection mechanism, making them less flexible when handling a large number of features.

\paragraph{Challenges in Selecting the Optimal MLP Configuration} The process of identifying the optimal MLP configuration is inherently time-consuming. 
We evaluated 36 distinct MLP configurations for each train-test pair by varying the number of hidden layers (1–4) and using the same number of neurons in each layer (2 to 512, in powers of 2). This comprehensive search is computationally intensive. Notably, we did not explore models with differing neuron counts across layers.

% \subsection*{Future Work} 
% Various neural network architectures can be explored. 
% Beyond MLPs, one might investigate the performance of other architectures tailored to the specific characteristics of the data. 
% In some cases, these alternative network structures may yield superior results compared to MLPs or linear models.
% About feature selection, instead of manual feature extraction, it could be worthwhile to explore Recurrent Neural Networks (RNNs) \citep{rnn} for directly extracting features from raw sequences. Variants like Gated Recurrent Units (GRUs) \citep{gru} or Long Short-Term Memory networks (LSTMs) \citep{lstm} could be examined for this task. Moreover, as feature transformations like logarithmic scaling were employed, it's worth exploring alternative approaches to feature engineering.

\section{Conclusion}\label{sec:conclusion}

The proposed model tends to achieve comparable or slightly higher accuracy and F1 scores in changepoint detection compared to linear and tree-based models.
Feature selection is crucial for MLP performance, as using all features doesn't necessarily improve performance.
The main limitation of the proposed model is the lack of an automatic feature selection mechanism, which makes it less robust than L1-regularized linear and tree-based models when dealing with a large set of features.

Based on the proposed model limitations, several potential research directions can be explored.
Regarding feature selection, rather than relying on manual feature extraction, exploring Recurrent Neural Networks (RNNs) \citep{rnn} - or its variants such as Gated Recurrent Units (GRUs) \citep{gru} and Long Short Term Memory networks (LSTMs) \citep{lstm} - for directly extracting features from raw sequences could be beneficial.

\subsection*{Reproducible Research Material}
For those interested in replicating our study, all the code and associated materials are available at this link: 
\href{https://github.com/lamtung16/ML_ChangepointDetection}{github.com/lamtung16/ML\textunderscore ChangepointDetection}.
This commitment to reproducibility ensures transparency and allows others to validate and build upon our findings.

\section*{Declarations}
No funding was received for conducting this study.

\newpage
\appendix
\section{Appendix}
\subsection{OP optimization problem} \label{OP optimization}
Find partition vector $\mathbf{m} \in \mathbb{R}^N$ that minimizes the following cost function for a given sequence $\mathbf{d} \in \mathbb{R}^N$ and $\lambda \geq 0$:
\begin{equation*} \label{eq:OP_loss}
C_\lambda(\textbf{d}, \textbf{m}) = \sum_{i=1}^N l(d_i, m_i) + \lambda I[m_i \neq m_{i+1}]
\end{equation*}

\noindent where $l(m_i, d_i)$ usually represents the squared error between the sequence value $d_i$ and the mean segment $m_i$, i.e., $l(d_i, m_i) = (m_i-d_i)^2$.
$I[m_i \neq m_{i+1}]$ is an indicator function that equals 1 if there is a changepoint (when $m_i \neq m_{i+1}$) and 0 otherwise.

\subsection{Supplemental Figure(s)} \label{sup_figures}

\begin{figure}[!h]
\centering
\includegraphics[width=0.99\textwidth]{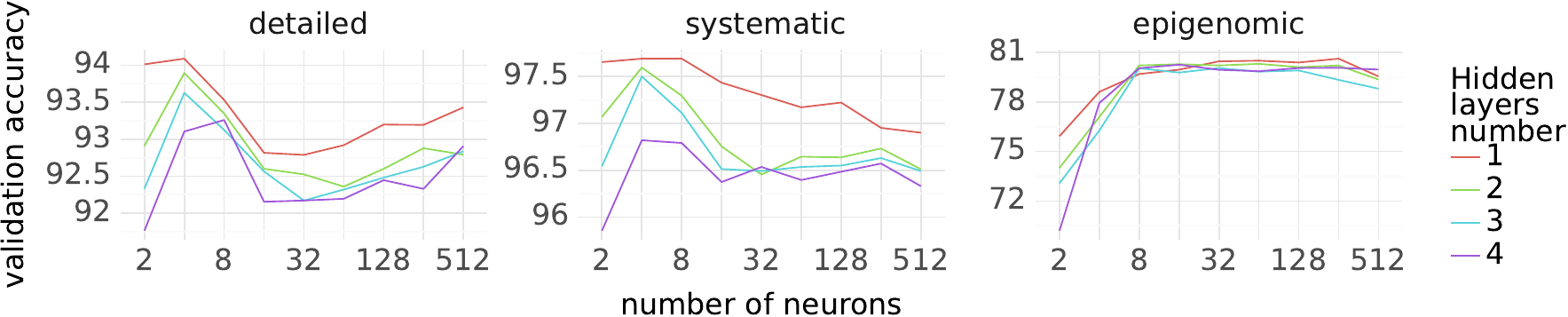}
\caption{MLP configurations' average validation accuracies, regardless of number of features.
For datasets detailed and systematic, it's generally preferable to have fewer hidden layers, typically around 4 to 8 neurons per layer suffice. In dataset epigenomics, where accuracy remains relatively consistent across various numbers of hidden layers, optimal accuracy tends to be achieved with a neuron count ranging from 8 to 256.}
\label{fig:mlp}
\end{figure}

\begin{figure}[!h]
\centering
\includegraphics[width=0.9\textwidth]{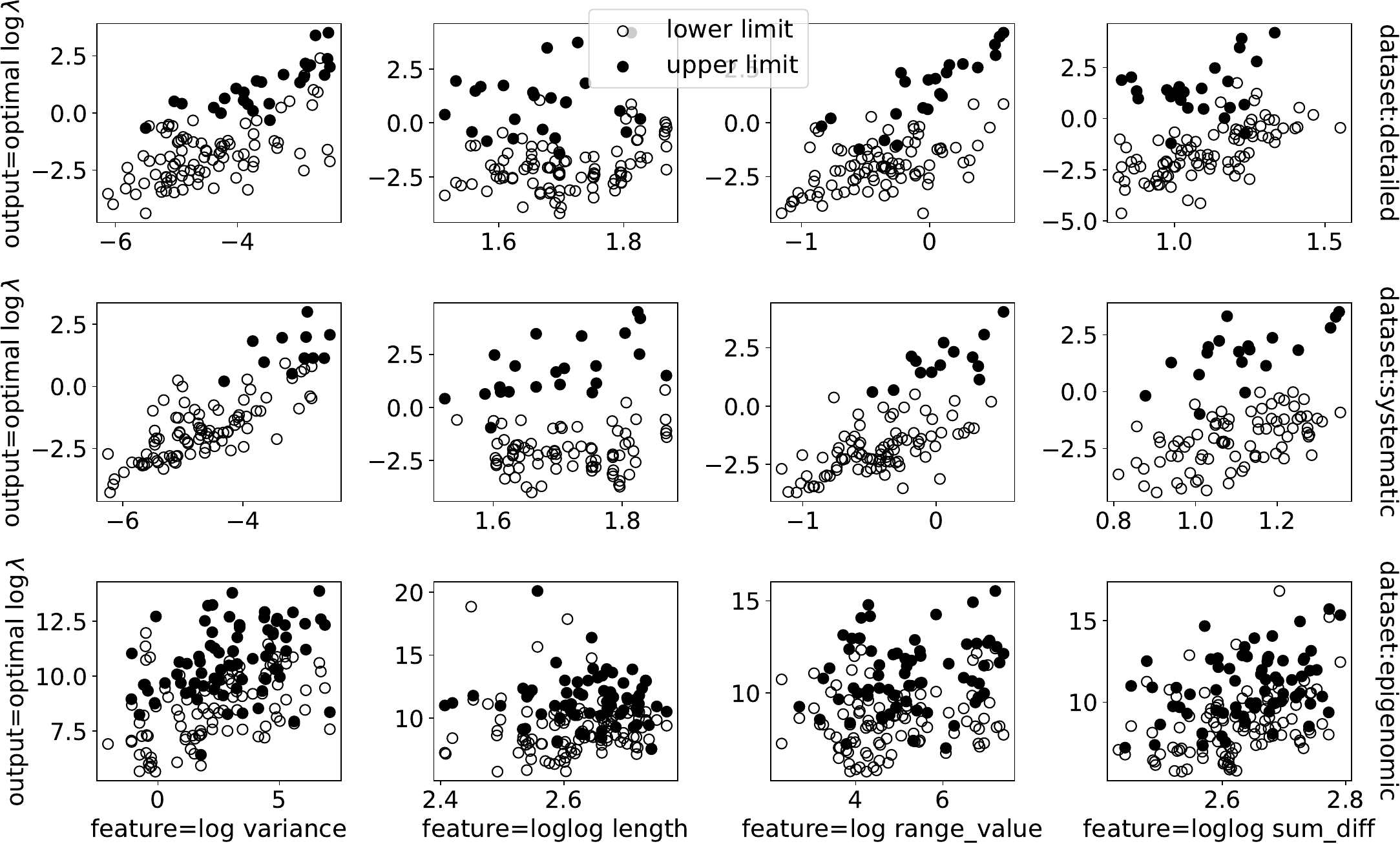}
\caption{Graphs illustrating the relationship between four features and target intervals. Variance and length are selected because previous studies have incorporated them into the model. Upon examination, we observe that the range value and sum of absolute differences exhibit a discernible increasing monotonic linear relationship with the target interval, indicating their potential as good features.}
\label{fig:features_targets}
\end{figure}

\newpage
\subsection{Supplemental Table(s)} \label{sup_tables}
\begin{table}[!h]
\centering
\caption{List of employed packages}\label{tab:packages}%
\begin{tabular}{@{}llll@{}}
\toprule
\textbf{Model} & \textbf{Package Source} & \textbf{Name} & \textbf{Prog. Language} \\
\midrule
Linear & \citep{penaltyLearning} & \texttt{penaltyLearning} & \texttt{R} \\
MMIT   & \citep{mmit_R} & \texttt{mmit} & \texttt{R} \\
AFT in XGBoost & \citep{xgboost_package} & \texttt{xgboost} & \texttt{Python} \\
MLP & \citep{pytorch} & \texttt{pytorch} & \texttt{Python} \\
\botrule
\end{tabular}
\end{table}

\clearpage
\bibliography{sn-bibliography}
\end{document}